\documentclass{article}

\usepackage{microtype}
\usepackage{graphicx}
\usepackage{subfigure}
\usepackage{booktabs} 
\usepackage{multirow}
\usepackage{float}
\usepackage{natbib}
\usepackage{hyperref}

\usepackage[accepted]{icml2020}

\icmltitlerunning{Interpretable EEG-to-Image Generation with Semantic Prompts}

\begin{document}

\twocolumn[
\icmltitle{Interpretable EEG-to-Image Generation with Semantic Prompts}



\icmlsetsymbol{equal}{*}

\begin{icmlauthorlist}
\icmlauthor{Arshak Rezvani}{equal,UBC1}
\icmlauthor{Ali Akbari}{equal,UBC1}
\icmlauthor{Kosar Sanjar Arani}{UT}
\icmlauthor{Maryam Mirian}{UBC1}
\icmlauthor{Emad Arasteh}{UBC1}
\icmlauthor{Martin J.~McKeown}{UBC1,UBC2}
\end{icmlauthorlist}

\icmlaffiliation{UBC1}{Pacific Parkinson’s Research Centre, Djavad Mowafaghian Centre for Brain Health, University of British Columbia, Vancouver, Canada}

\icmlaffiliation{UT}{Department of Electrical and Computer Engineering, University of Tehran, Tehran, Iran}

\icmlaffiliation{UBC2}{Department of Medicine
British Columbia, Vancouver, Canada}

\icmlcorrespondingauthor{Arshak Rezvani}{arshak.rezvani0@gmail.com}
\icmlcorrespondingauthor{Emad Arasteh}{emad.arasteh@ubc.ca}

\icmlkeywords{Machine Learning, ICML}

\vskip 0.3in
]



\printAffiliationsAndNotice{\icmlEqualContribution} 

\begin{abstract}
Decoding visual experience from brain signals offers exciting possibilities for neuroscience and interpretable AI. While EEG is accessible and temporally precise, its limitations in spatial detail hinder image reconstruction. Our model bypasses direct EEG-to-image generation by aligning EEG signals with multilevel semantic captions—ranging from object-level to abstract themes—generated by a large language model. A transformer-based EEG encoder maps brain activity to these captions through contrastive learning. During inference, caption embeddings retrieved via projection heads condition a pretrained latent diffusion model for image generation. This text-mediated framework yields state-of-the-art visual decoding on EEGCVPR dataset, with interpretable alignment to known neurocognitive pathways. Dominant EEG-caption associations reflected importance of different semantic levels extracted from perceived images. Saliency maps and t-SNE projections reveal semantic topography across the scalp. Our model demonstrates how structured semantic mediation enables cognitively-aligned visual decoding from EEG.
\end{abstract}

\section{Introduction}
\label{Introduction}
Decoding visual perception from brain activity supports a range of applications, including brain-computer interfaces, neurofeedback, and cognitive monitoring \cite{decoding2024}. While fMRI enables detailed visual reconstructions, its cost and lack of real-time viability have motivated interest in EEG-based methods \cite{eeg2023}. EEG offers high temporal resolution and low cost, but its noisy and coarse spatial characteristics limit reconstruction quality \cite{visualstimuli}.

Recent advances leverage pretrained generative models, particularly diffusion models, to enhance EEG-driven image synthesis \cite{natural2023, li2024visualdecoding}. Contrastive learning further strengthens this process by embedding EEG and visual representations into a shared latent space \cite{song2024eeg2image, xu2024alljoined1}. Methods like CLIP have shown potential for cross-modal alignment and have been adapted for EEG-based joint representation learning \cite{akbari2024hierarchicalvit, singh2024learning}. However, reconstructions often reflect biases or hallucinations rather than the perceptual authenticity of the reference image\cite{shirakawa2025spuriousreconstructionbrainactivity}. While learning semantic similarity has improved image robustness \cite{zhang2023mmcl}, the role of semantic structure in EEG-image decoding remains underexplored.

Li et al. \cite{li2024visualdecoding} introduced image-derived text to bridge EEG and image modalities, but prioritized EEG-image alignment over EEG-text understanding. Moreover, their reliance on trial-averaged EEG and dataset-specific tuning \cite{hebart2019things} limits real-world applicability.

We address these gaps through a model that aligns EEG with multilevel semantic prompts—including object, spatial, and thematic content—generated via LLMs. These captions are retrieved using a multi-head transformer encoder trained with contrastive loss, and then passed to a frozen, pretrained diffusion model \cite{sauer2024adversarial}. This enables interpretable, scalable decoding across varied perceptual content.

Our main contributions are as follows. First, we introduce multilevel semantic prompts, where large language model (LLM)-generated captions span from object-level descriptions to abstract visual themes. Second, we present a text-to-image synthesis framework in which EEG-retrieved captions are used to guide a frozen diffusion model for image generation. Third, we propose a semantic-aligned neural mapping approach that employs a multi-head encoder to associate the importance of EEG channels with distinct layers of visual semantics.

\section{Preliminaries and Related Works}
\label{gen_inst}
Early EEG-to-image reconstruction models, such as Brain2Image \cite{kavasidis2017brain2image} and GAN-based approaches by Palazzo et al. \cite{palazzo2017ganbrain}, showed limited visual fidelity, with Inception Scores (IS) of 4.49 and 5.07, respectively. Spampinato et al. \cite{spampinato2019deeplearning} focused on classification using LSTM encoders, achieving 82.9\% accuracy on EEGCVPR, but did not attempt reconstruction.

Akbari et al. \cite{akbari2024hierarchicalvit} advanced the field with Hierarchical-ViT and StyleGAN, achieving improved metrics on EEGCVPR (IS: 12.17, FID: 122.91, KID: 0.059). More recently, the use of latent diffusion models pretrained on large-scale text-image datasets such as LAION-5B \cite{rombach2022ldm}, and contrastive methods like CLIP \cite{radford2021learning}, have significantly boosted reconstruction quality \cite{li2024visualdecoding, fei2025perceptogramreconstructingvisualpercepts, bai2025dreamdiffusion, fu2025brainvis}.

LLMs like GPT-4 \cite{openai2024gpt4} have also enhanced visual captioning, supporting semantic understanding in generative pipelines \cite{bonillasalvador2024pixloredatasetdrivenapproachrich}. Li et al. \cite{li2024visualdecoding} proposed EEG-guided image generation with a text mediator but relied on a single caption per image, limiting semantic depth.

In summary, most prior methods either use minimal semantic text input or lack a principled way to align EEG with visual meaning. This restricts both the realism and interpretability of the output, underscoring the need for deeper semantic grounding in EEG-to-image frameworks.

\section{Method}
\label{headings}
\begin{figure*}[ht!]
	\centering
\includegraphics[width=\linewidth]{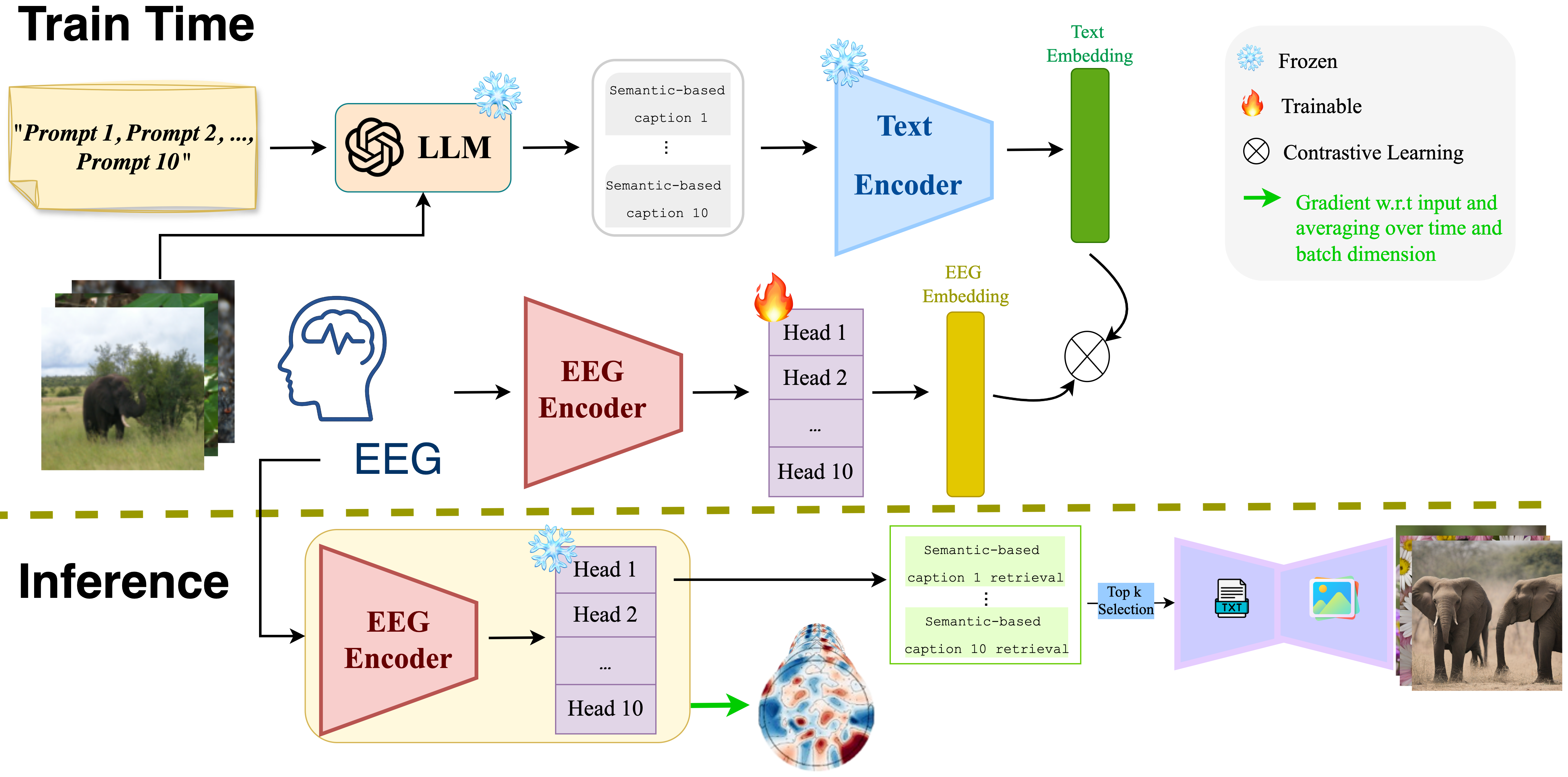}	\caption{Framework overview. (Top) Training: EEG aligned with semantic captions via contrastive learning. (Bottom) Inference: Retrieved captions condition a pretrained diffusion model.}
	\label{fig:fig1}
\end{figure*}
\subsection{Motivation and Overview}
Pretrained latent diffusion models such as Stable Diffusion \cite{rombach2022ldm} offer strong generative capabilities, applied to extensive image-text datasets. By aligning EEG to CLIP’s embedding space \cite{radford2021clip}, we leverage transfer learning to enhance image reconstruction from limited data. LLMs (e.g., GPT-4 \href{https://openai.com/research/gpt-4}{[GPT-4]}) are used to extract multilevel captions reflecting object, spatial, and thematic features. We assess image quality using IS and KID, and similarity via features from AlexNet, Inception, CLIP, SwAV, SSIM, and PixCorr. Fig.~\ref{fig:fig1} illustrates our architecture.
\subsection{Training Phase}
Each image is described with 10 captions generated using an LLM, grouped into three semantic levels. Low-level prompts target object and color attributes \cite{xu2024deepimage}, mid-level prompts capture scene and spatial layout \cite{walther2023midlevel, kubilius2014midlevel}, and high-level prompts focus on emotion and theme \cite{park2025hierarchical, pandiani2024abstract}. These structured captions form the bridge between EEG signals and text embeddings.
We used a transformer-based EEG encoder to capture spatiotemporal dependencies in EEG \cite{pfeffer2024transformer}, applying spatial and temporal attention to model dynamics beyond CNN-based approaches \cite{du2022eeg}. 
The EEG encoder is trained with a CLIP-style contrastive loss \cite{radford2021clip}, aligning EEG projections with the corresponding caption embeddings while repelling non-matching pairs. 
\subsection{Inference Phase}
At test time, EEG signals are mapped into the CLIP space via the encoder. Each head retrieves the most similar captions using cosine similarity, which are concatenated to condition a frozen diffusion model \cite{rombach2022ldm}. This enables text-to-image synthesis using pretrained text-image alignment.
\subsection{Saliency-Based Interpretability}
To assess the sensitivity of the model's predictions to the input EEG signal, we computed the gradient of the loss related to the input. This procedure identified which spatiotemporal components of the EEG contribute most strongly to the model's output. Specifically, we enabled gradient tracking on the input EEG batch and performed a forward pass through the frozen model to compute the contrastive loss between the EEG and corresponding text embeddings. We then computed backpropagation loss to obtain the gradient with respect to the input EEG signal. These gradients were aggregated over the batch dimension, and the resulting saliency maps are averaged across the time dimension to obtain per-channel importance scores.
\subsection{Ablation Studies}
We evaluated the role of contrastive loss vs. MSE and tested multilevel captions against single-caption setups. Comparisons were made using EEG-caption accuracy and downstream image quality metrics.

\section{Experimental Results}

\begin{figure}[ht!]
    \centering
    \subfigure[Real Images]{
        \begin{minipage}[t]{0.45\textwidth}
            \centering
            \includegraphics[width=0.18\textwidth]{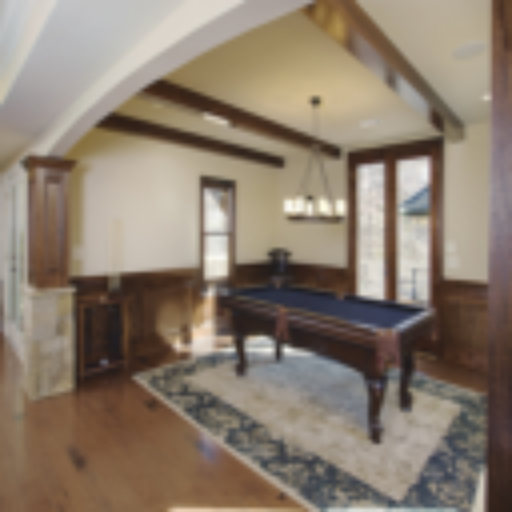}
            \includegraphics[width=0.18\textwidth]{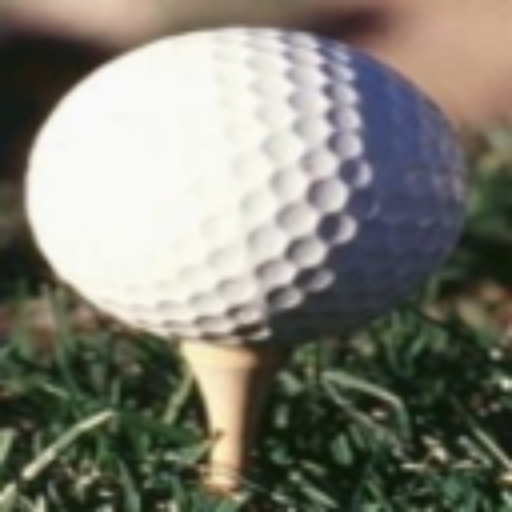}
            \includegraphics[width=0.18\textwidth]{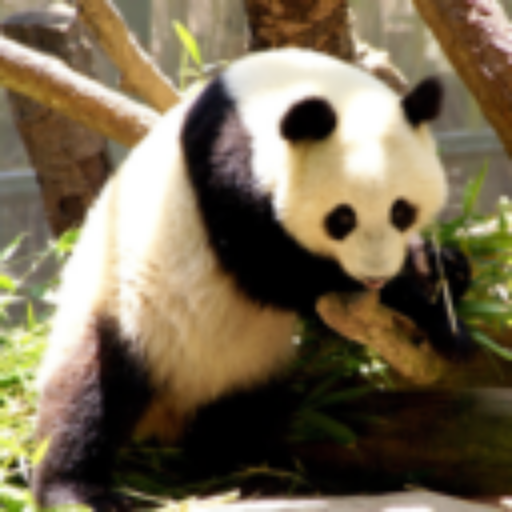}
            \includegraphics[width=0.18\textwidth]{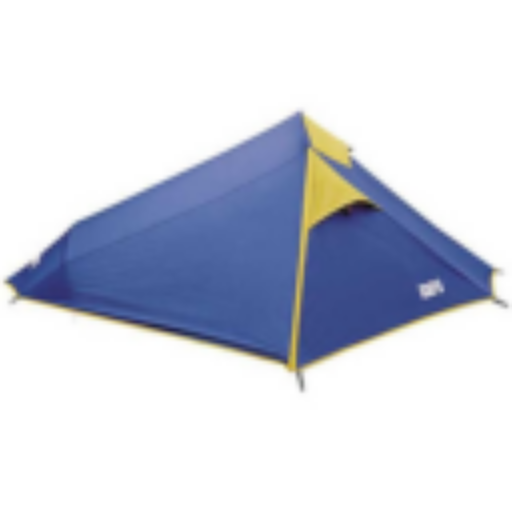}
            \includegraphics[width=0.18\textwidth]{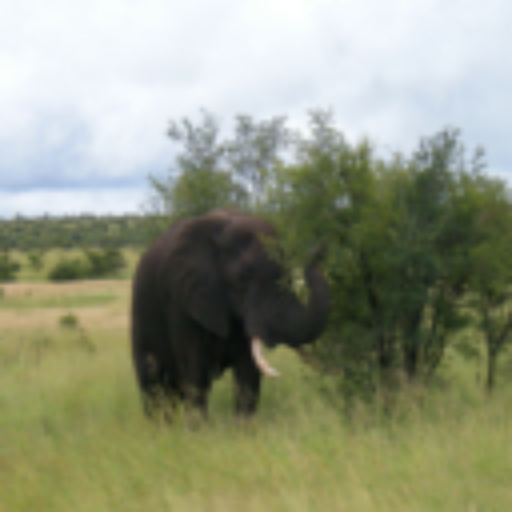} \\
            \includegraphics[width=0.18\textwidth]{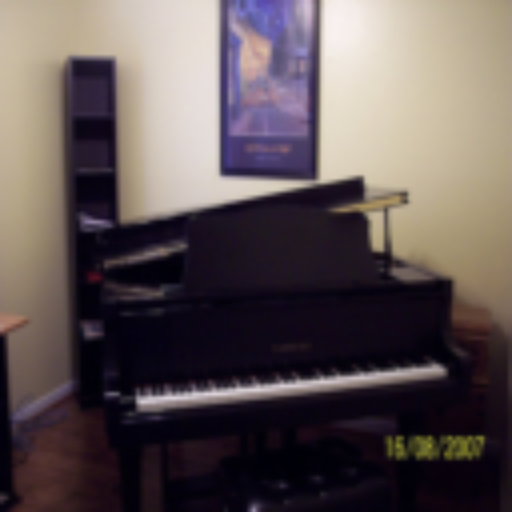}
            \includegraphics[width=0.18\textwidth]{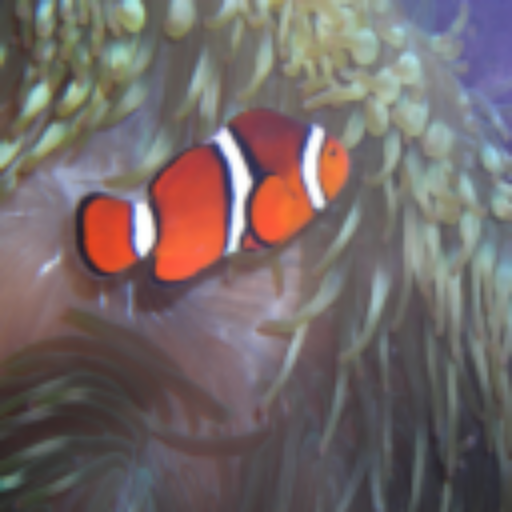}
            \includegraphics[width=0.18\textwidth]{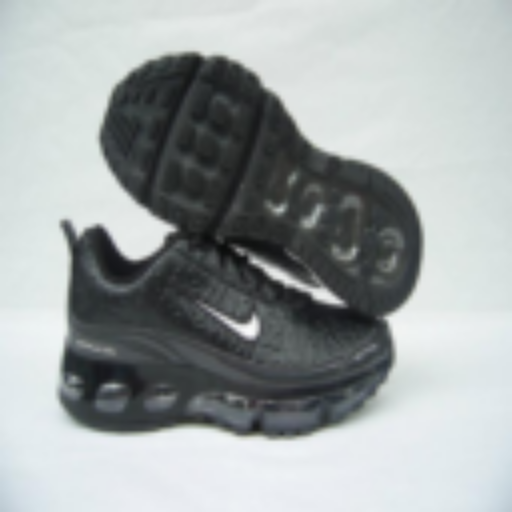}
            \includegraphics[width=0.18\textwidth]{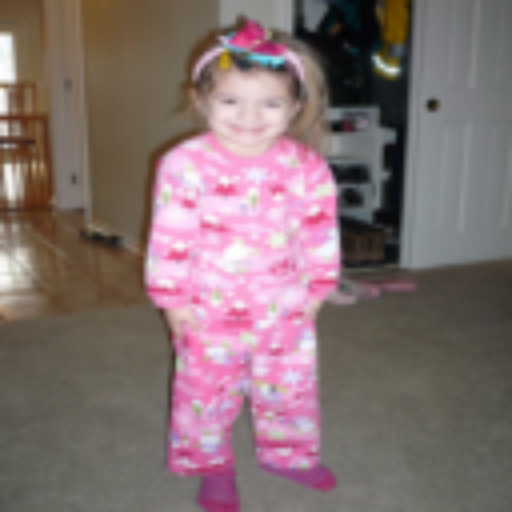}
            \includegraphics[width=0.18\textwidth]{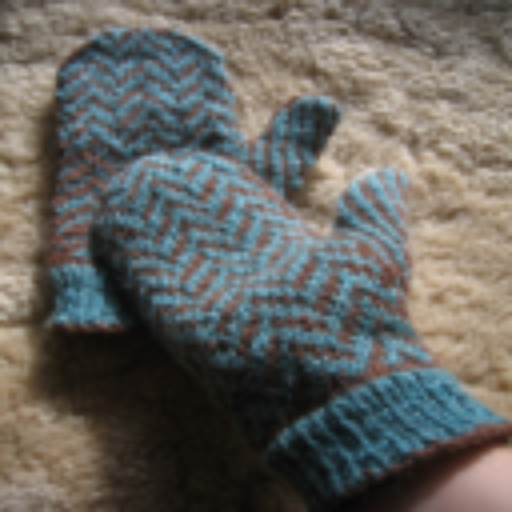}
        \end{minipage}
    }
    \hspace{0.5cm}
    \subfigure[Generated Images]{
        \begin{minipage}[t]{0.45\textwidth}
            \centering
            \includegraphics[width=0.18\textwidth]{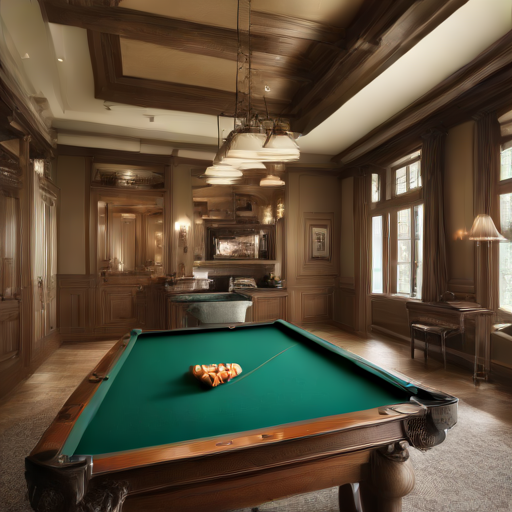}
            \includegraphics[width=0.18\textwidth]{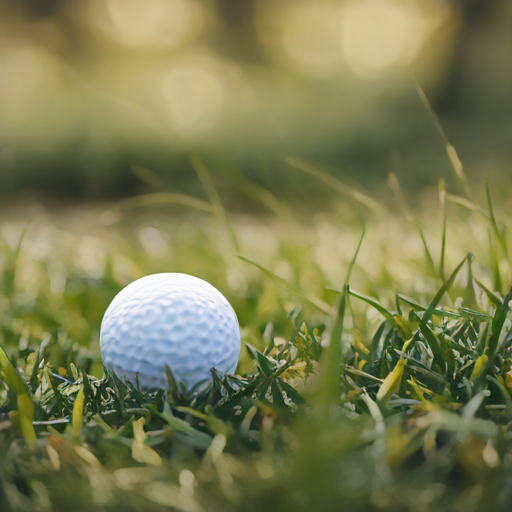}
            \includegraphics[width=0.18\textwidth]{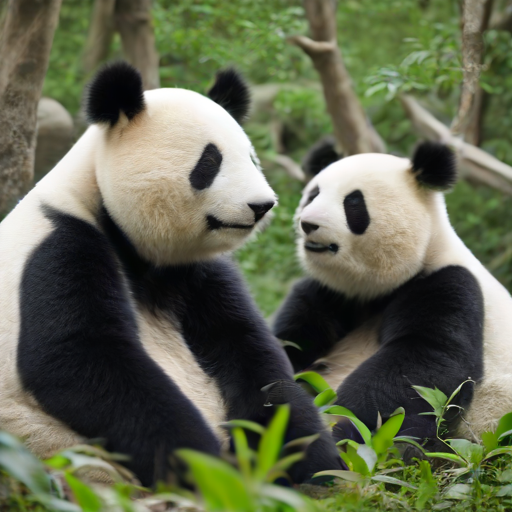}
            \includegraphics[width=0.18\textwidth]{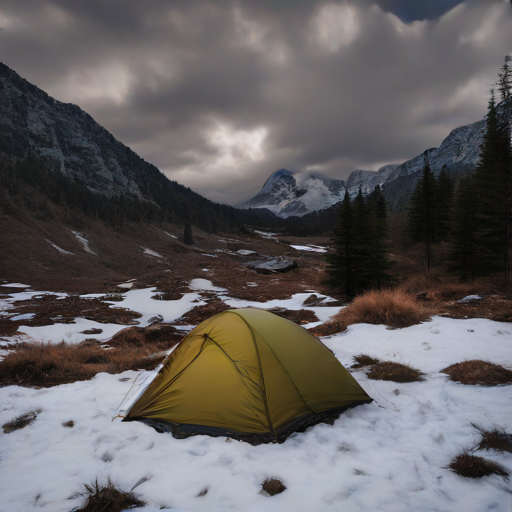}
            \includegraphics[width=0.18\textwidth]{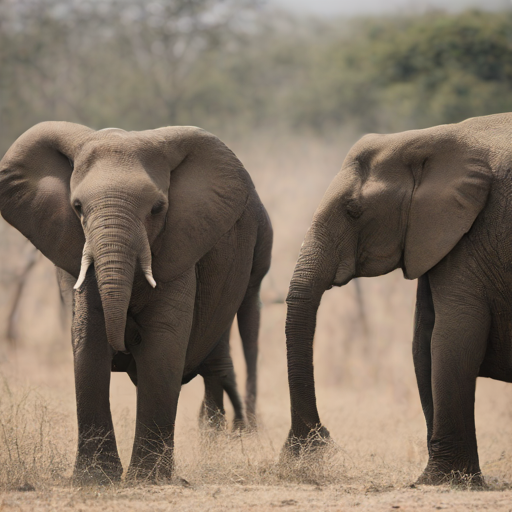} \\
            \includegraphics[width=0.18\textwidth]{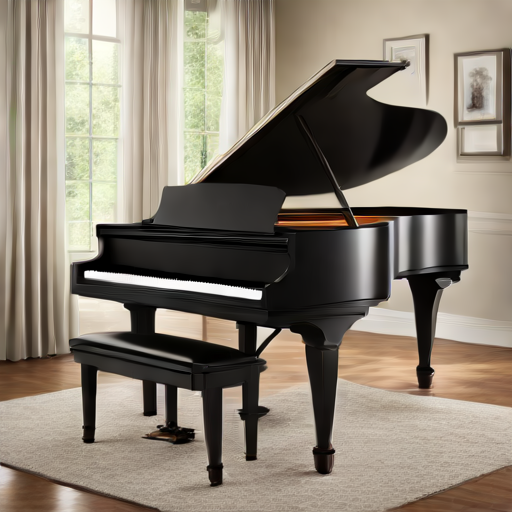}
            \includegraphics[width=0.18\textwidth]{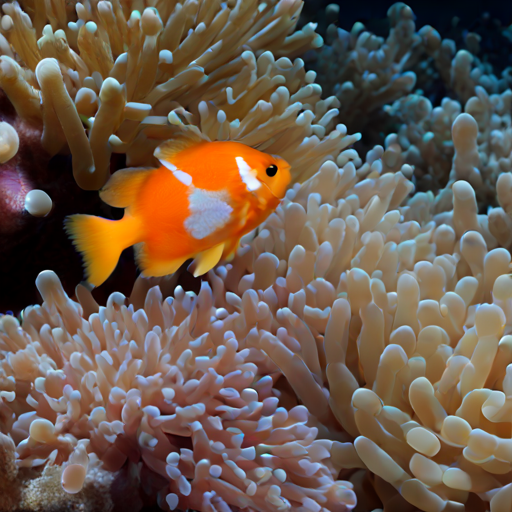}
            \includegraphics[width=0.18\textwidth]{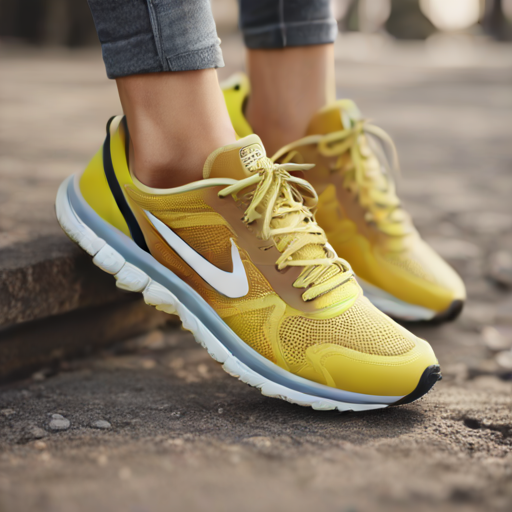}
            \includegraphics[width=0.18\textwidth]{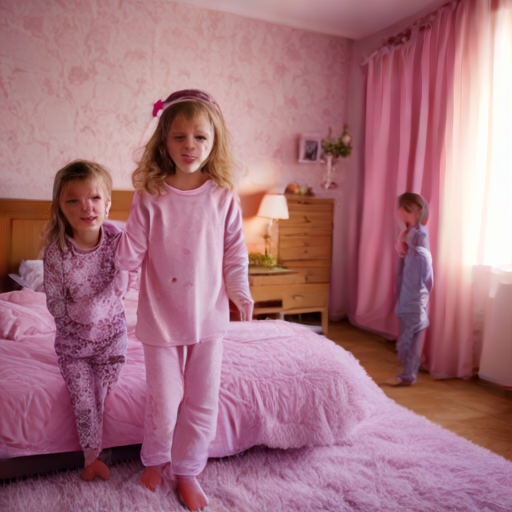}
            \includegraphics[width=0.18\textwidth]{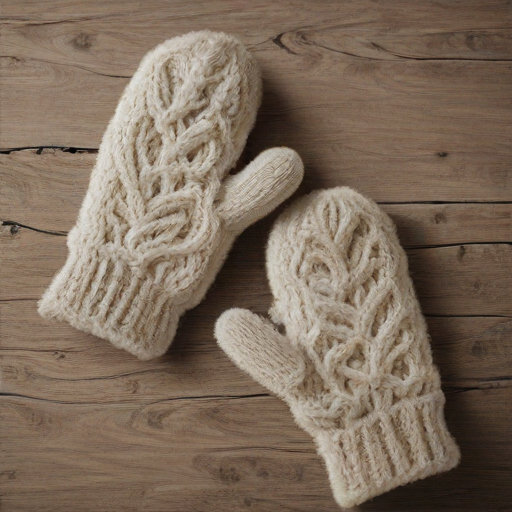}
        \end{minipage}
    }
    \vspace{0.2cm} 
    \caption{Comparison of groundtruth and generated images for the EEGCVPR dataset.}
    \label{fig:generated}
    \vspace{-0.2cm}
\end{figure}

We evaluated our model on the EEGCVPR dataset \cite{singh2024learning}, which includes EEG recordings from six subjects across 40 object categories using a 128-channel cap at 1,000 Hz. We assessed EEG-text-image alignment, classification, saliency patterns, head specialization, and image generation quality. Using an ensemble of heads \cite{jurek2014survey}, classification accuracy on EEGCVPR reaches 79\%. Our model achieves state-of-the-art image generation results on EEGCVPR across multiple metrics, including IS, KID, AlexNet (2/5), Inception, CLIP Score, SwAV, PixCorr, and SSIM (see Table~\ref{table1}). FID is reported (64.2) but not emphasized due to known limitations \cite{jeevan2024fld}. Visual examples are shown in Fig.~\ref{fig:generated}.
\begin{table*}[ht!] 
	\caption{Comparison of our model with state-of-the-art EEG-to-image reconstruction methods on EEGCVPR dataset.}
	\label{table1}
	\vspace{10pt} 
	\centering
	{\renewcommand{\arraystretch}{1.6} 
		\large 
		\resizebox{\textwidth}{!}{%
			\begin{tabular}{llccccccccccc}
				\toprule
				\textbf{Dataset} & \textbf{Model} & \textbf{Type} & \textbf{IS$\uparrow$} & \textbf{KID$\downarrow$}  &
				\textbf{PixCorr$\uparrow$} &
				\textbf{SSIM$\uparrow$} & \textbf{Alex2$\uparrow$} & \textbf{Alex5$\uparrow$} & \textbf{Inception$\uparrow$} & \textbf{CS$\uparrow$} & \textbf{SwAV$\downarrow$} \\
				\midrule
				\multirow{6}{*}{EEGCVPR \cite{8099962}} & 
				EEGStyleGAN-ADA \cite{singh2024learning} & GAN & 10.82 & 0.56 & - & - & - & - & - & - & - \\
				& EEG-ViT \cite{akbari2024hierarchicalvit} & GAN & 12.17 & 0.05 & - & - & - & - & - & - & - \\
				& NeuroVision \cite{khare2022neurovision} & GAN & 5.15 & - & - & - & - & - & - & - & - \\
				& Improved-SNGAN \cite{zheng2020decoding} & GAN & 5.53 & - & - & - & - & - & - & - & - \\
				& Brain2Image-VAE \cite{kavasidis2017brain2image} & VAE & 4.49 & - & - & - & - & - & - & - & - \\
				& \textbf{Ours} & Diffusion & \textbf{37.29 ± 0.32} & \textbf{0.009 ± 0.009} & \textbf{0.06} & \textbf{0.30} & \textbf{0.65} & \textbf{0.80} & \textbf{0.88} & \textbf{0.88} & \textbf{0.57} \\
				\bottomrule
			\end{tabular}
	}}
\end{table*}

\subsection{t-SNE Visualization of Semantic Projections}
\begin{figure}[ht!]
\begin{center}
\centerline{\includegraphics[width=\linewidth, height=7cm]{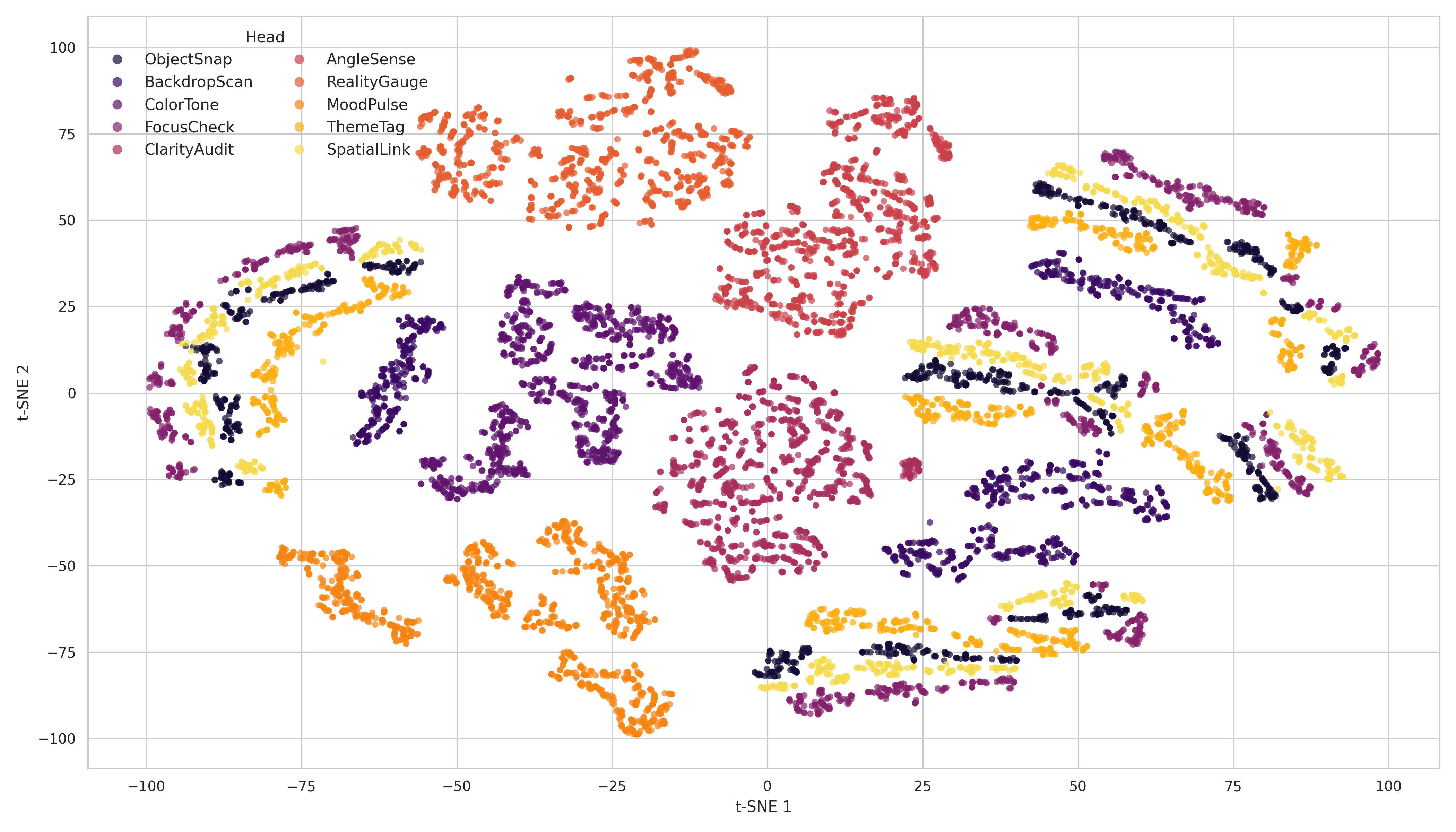}} %
\caption{\small t-SNE of EEG encoder outputs, colored by projection head (EEGCVPR).}
\label{fig:tsne_all_heads}
\end{center}
\vskip -0.2in
\end{figure}

To examine semantic separability across encoder heads, we used t-SNE to visualize EEG embeddings. As shown in Fig.~\ref{fig:tsne_all_heads}, each head yields a distinct representation cluster, supporting the specialization of heads for different semantic levels.
\vspace{-1ex}
\subsection{Topographic Saliency Maps Across Semantic Levels}
\begin{figure}[ht]
\vskip 0.2in
\begin{center}
\centerline{\includegraphics[width=\columnwidth]{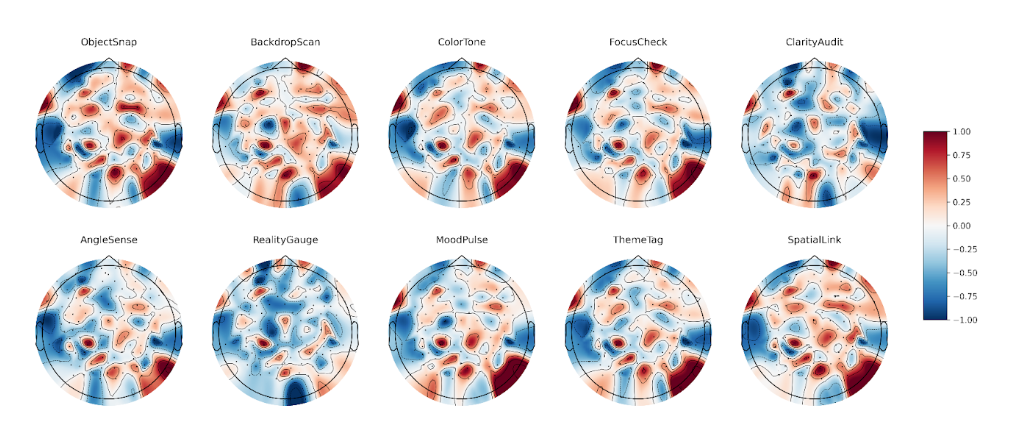}}
\caption{Topographic saliency maps showing channel specialization across semantic heads.}
\label{fig:Topographic}
\end{center}
\vskip -0.2in
\end{figure}

We used gradient-based saliency to map EEG channel importance across semantic levels. As shown in Fig.~\ref{fig:Topographic}, topographic patterns linking EEG activity to semantic dimensions are observed. Low-level features (e.g., color, clarity) evoke activation over occipital electrodes (e.g., O1, Oz), while high-level semantics (e.g., mood, theme) engage more frontal regions (e.g., Fz, FC1), in line with visual and executive processing \cite{portin1998neuromagnetic, garcia2022semantic}. Spatially structured concepts (e.g., layout, angle) show parietal localization \cite{hnazaee2018semantic}
\subsection{Head-Level Semantic Specialization}
\begin{figure}[ht!]
\begin{center}
\centerline{\includegraphics[width=\columnwidth]{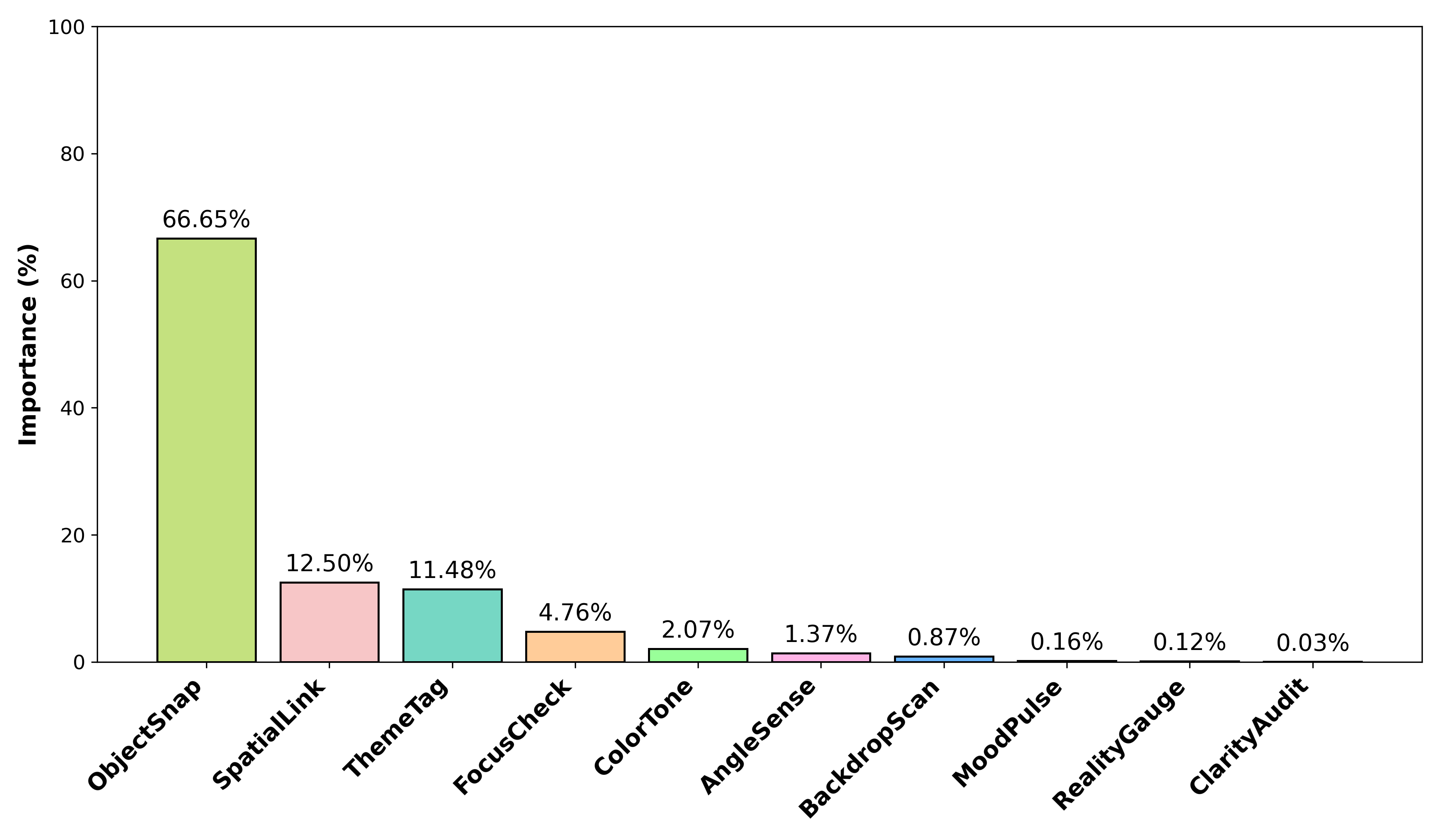}}
\caption{Distribution of top-caption matches across EEG encoder heads (for the EEGCVPR dataset).}
\label{fig:pie_plot}
\end{center}
\vskip -0.2in
\end{figure}

To quantify the role of each semantic level, we computed how often each EEG encoder head produced the top-caption match during retrieval. As shown in Fig. \ref{fig:pie_plot}, majority of the alignment instances were concentrated in ObjectSnap, SpatialLink, and ThemeTag. 
\vspace{-0.5\baselineskip} 
\subsection{Ablation Studies}
We compare CLIP-based vs. MSE loss and assess semantic head combinations. CLIP loss yields higher accuracy in most heads, validating its role in semantic alignment. Using top-2 heads (IS = 38.00, KID = 0.0106) outperforms other configurations. 

\section{Discussion}
\label{sec:Discussion}
Our results show that EEG serves as a signal aligned with structured semantic concepts. By leveraging multilevel captions as an interpretive bridge, our model enables high-fidelity reconstruction via pretrained diffusion models, while enhancing semantic clarity. t-SNE visualizations and head-wise clustering indicate that EEG encoder heads specialize in distinct semantic levels. Ablation studies confirm that multi-head contrastive alignment outperforms single-caption setups, supporting semantic disentanglement.

Topographic saliency maps (Fig.~\ref{fig:Topographic}) validate known neurocognitive mappings: occipital channels respond to low-level features (e.g., color, clarity) \cite{portin1998neuromagnetic}, while frontal areas correspond to higher-level constructs (e.g., emotion, themes) \cite{garcia2022semantic}. ObjectSnap, SpatialLink, and ThemeTag emerge as the dominant heads (Fig.~\ref{fig:pie_plot}), reinforcing theories of distributed visual perception \cite{epstein2019scene, woods2001auditory, bar2006topdown}. Limitations include reliance on LLM-generated captions, a single tested dataset, and potential subject variability. 
\section{Conclusion}

We presented a novel EEG-to-image reconstruction framework that aligns multilevel semantic prompts with neural signals to guide diffusion-based generation. Our model achieves interpretable and high-quality visual decoding by integrating semantic specialization and transfer learning. Future work will explore refined caption selection and deeper analysis of EEG-semantic mappings to further improve resolution and generalization.

\bibliographystyle{icml2020}
\bibliography{references}

\end{document}